\documentclass[runningheads]{llncs}

\usepackage{eccv}
\usepackage{graphicx}
\usepackage{booktabs}
\usepackage[accsupp]{axessibility}
\usepackage{amsmath}
\usepackage{array}
\usepackage{xcolor}
\usepackage{colortbl}
\usepackage{tabularx}
\usepackage{multirow}
\usepackage[skip=8pt]{caption}
\usepackage[breaklinks,colorlinks,citecolor=blue]{hyperref}
\usepackage{orcidlink}

\newcolumntype{C}[1]{>{\centering\arraybackslash}p{#1}}
\newcommand{\g}[1]{\textcolor{gray}{#1}}

\begin{document}

\title{Understanding Why Foundation Models Work for Diffusion-Generated Image Detection}
\titlerunning{Understanding Why Foundation Models Work}

\author{
Davide Cozzolino \orcidlink{0000-0001-6158-7595} \and 
Giovanni Poggi \orcidlink{0000-0003-1327-4812} \and
Luisa Verdoliva \orcidlink{0000-0001-7286-7963}
}
\authorrunning{Cozzolino et al.}

\institute{University Federico II of Naples, 80125 Naples, Italy
\email{\{davide.cozzolino,poggi,verdoliv\}@unina.it}}

\maketitle

\begin{abstract}
Vision foundation models have recently emerged as powerful feature extractors for detecting AI-generated images, achieving strong generalization across generators and robustness to common image degradations.   However, the reason behind their effectiveness is poorly understood. In this work, we investigate what cues are exploited by foundation-model-based detectors to distinguish real images from diffusion-generated ones. 
To this end, we design an ad hoc analysis protocol based on DDIM inversion. Given a real image we generate a sequence of synthetic copies by changing the depth of DDIM inversion. Even though most copies are semantically identical to the real reference, the detector score varies significantly across them due to subtle traces introduced by the diffusion synthesis, showing that its decision is not primarily driven by semantic failures. 
Through a frequency-swapping analysis, we further reveal that the discriminative cues exploited by the detectors are mainly localized in the low-to-mid frequency range, rather than only in the high-frequency range, as is the case for
artifacts commonly associated with generative models. Finally, a latent-space analysis shows that regenerated images exhibit reduced variance and effective dimensionality, indicating that diffusion models do not fully reproduce the variability of real data. Overall, our results suggest that foundation-model-based detectors succeed by capturing non-semantic low-to-mid frequency distributional discrepancies between real and diffusion-generated images. These findings provide new insight into the robustness and generalization of such detectors and suggest directions for more interpretable forensic methods.
\end{abstract}

\section{Introduction}

Detecting whether an image is real or synthetically generated has become increasingly important, as modern generative models can produce highly realistic visual content that may be misused for disinformation \cite{barrett2024identifying,bontcheva2024generative}. Accordingly, a large body of work has focused on designing detectors capable of distinguishing real from generated images \cite{lin2024detecting,amerini2025deepfake}.
Beyond obvious visual failures, such as clear asymmetries, inconsistent perspective or implausible object geometry, that can be analyzed by vision-language models \cite{tan2026forendex,sun2026salartvqa}, synthetic images often contain subtle statistical traces. As generative models produce increasingly realistic images, such hidden artifacts become more relevant to design effective detectors. 

\begin{figure}[t!]
    \centering
    \includegraphics[width=1.0\linewidth,page=1,clip,trim=0 90 0 0]{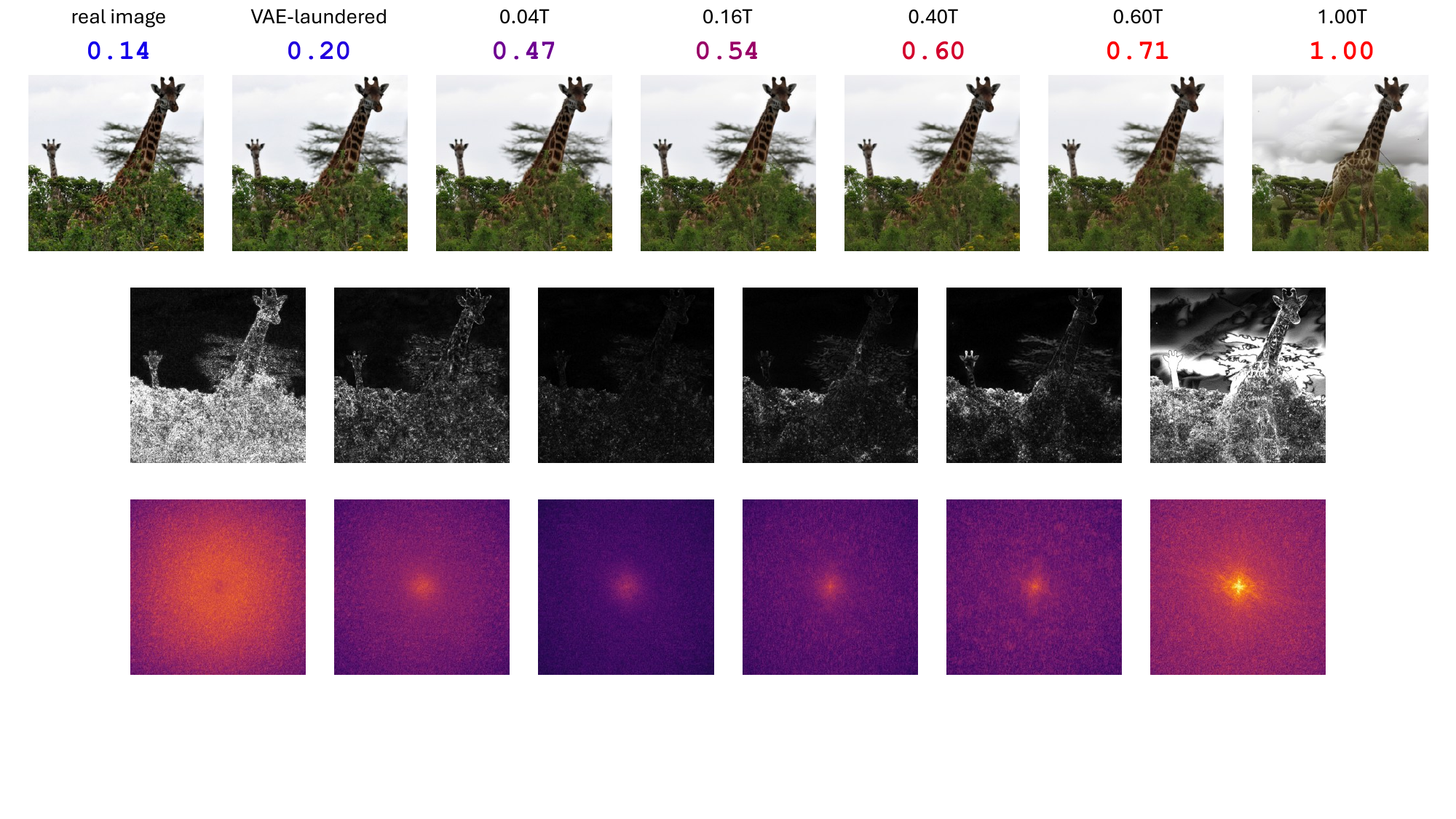}
        \caption{
        Visualization of diffusion-induced traces across regeneration timesteps. The first row shows the original real image, its VAE-laundered version, and DDIM-regenerated images obtained from increasing inversion timesteps $k$, together with the corresponding detector scores. The second row reports the pixel-wise difference with respect to the previous image, while the third row shows the corresponding power spectra. As $k$ increases, the detector score progressively shifts toward the fake class, and the difference maps reveal increasingly structured artifacts. Notably, these traces are visible not only at high frequencies but especially in the low-frequency region of the spectrum.
} 
    \label{fig:teaser}
\end{figure}

Early works have shown that synthetic images exhibit telltale artifacts in the Fourier domain \cite{zhang2019detecting} such as pronounced peaks at harmonic frequencies and anomalous shapes of the power spectrum.
These artifacts are typically referred to as low-level traces, since they are not related to the semantic content of the image but arise from the image generative process itself.
For example, in generative adversarial networks, spectral peaks are due to common architectural components that perform the necessary upsampling operations \cite{Frank2020frequency,Durall2020upconv}.
Likewise, similar artifacts appear in images generated by diffusion models (DM), introduced in the final decoding process from the latent space.
In addition, power spectrum anomalies have been also observed at the medium and high frequencies, especially along diagonal directions \cite{corvi2023intriguing}.
These are precious cues for discrimination, largely exploited, either explicitly or implicitly, by a plethora of dedicated models proposed in recent years for AI-generated content detection \cite{li2024masksim,Karageorgiou2025spai,corvi2025seeing,li2026spectral,gong2026cross}.
However, relying on high-frequency artifacts is a double-edged sword: it yields strong performance under ideal conditions but leads to a drastic drop in performance when these traces are attenuated by low-pass filtering. This often occurs in real-world settings, where images shared on social networks are resized and compressed, removing many low-level cues. 
As a result, detectors suffer from poor robustness and limited generalization, making them unreliable in the wild \cite{tariang2024synthetic}.

In this challenging scenario, detectors based on foundation models stand out as a notable exception.
Recent studies have shown that frozen pre-trained large Vision Foundation Models (VFMs) provide remarkably effective and robust representations for detecting AI-generated images.
A simple linear classifier working on features extracted from foundation models provides excellent results and strong generalization across different generative models, often outperforming detectors specifically trained for the real-versus-fake task \cite{ojha2023towards}.
Furthermore, this simple detector proves robust to common low-level perturbations and even laundering attacks, suggesting that it does not rely primarily on spectral peaks or high-frequency artifacts, but that alternative discriminating features come into play.
This was explicitly shown in \cite{cozzolino2024raising}, where CLIP features were found to be largely independent of, and complementary to, low-level forensic traces.
Further evidence comes from experiments on more recent models, such as MetaCLIP 2 and DINOv3, which show better performance and greater robustness than previous models, even on more challenging in-the-wild datasets \cite{zhou2026simplicity,huang2025rethinking}.
Nevertheless, the reasons underlying this success are not yet fully understood. 
Shedding light on this mechanism is the primary goal of this paper.

We will show that diffusion models introduce artifacts at the medium and low frequencies of the spectrum, besides those already well-known at the high frequencies.
Furthermore, we show that foundation model-based detectors rely on these artifacts to make their decision.
To support this conclusion, we design a testing strategy based on DDIM inversion.
Given a real image, we apply DDIM inversion \cite{song2020denoising} to derive a noise-like seed that
once fed to the DDIM generation process, allows us to re-generate an image very similar to the original image.
In this way, we obtain two images, one real and one generated by a diffusion model, which are semantically identical, but whose different origins are not lost to the detector.
Continuing along this path, we run the DDIM generation multiple times, starting form different timesteps,
thus generating a {\em sequence} of semantically coherent images that change very gradually along the process, which allows us to study in detail when and where artifacts appear and how they impact on the detector decision.
Fig.~\ref{fig:teaser} shows an example subsequence.
The rightmost image is generated with full DDIM inversion (1.00$T$) 
and is the only one that, due to process imperfections, differs semantically from the real image
(the leftmost one).
With partial inversion ($\leq$0.60$T$) semantic consistency is always ensured, yet the detector provides high (fake) scores down to values close to 0 ($\leq$0.04$T$).
The difference images show that artifacts appear throughout the entire process, not just at the end, and the corresponding spectra prove that they are mainly localized at low frequencies.

Overall, this work provides new insights into the behavior of foundation-model-based detectors and makes the following contributions:
\begin{itemize}
\item   We design a testing strategy based on DDIM inversion to expose the non-semantic artifacts introduced by diffusion models and arguably exploited by foundation models to detect synthetic images;
\item   We reveal a clear distributional gap between real and DM-generated images, with the most relevant differences emerging at mid and low spectral frequencies;
\item   We introduce several latent-space measures to quantify experimentally the inconsistencies observed in diffusion-based generative models.
\end{itemize}

\section{Related Work}

\noindent
{\bf Forensic detectors based on large VFMs.}
The advent of large language and vision models, with their unprecedented representation ability, has stimulated intense research in the computer vision community, including image forensics, where they have demonstrated strong potential \cite{sha2022fake,ojha2023towards,cozzolino2024raising}.
CLIP features have proven highly effective for distinguishing real images from synthetic ones, even based on a simple nearest-neighbor or linear classifier \cite{ojha2023towards}, and even when trained on just a few images \cite{cozzolino2024raising}.
These properties hold even more true for recent pretrained large models, such as MetaCLIP2, DINOv3, PE-Core \cite{huang2025rethinking,zhou2026simplicity,lee2026ssafe,choi2026anchor}.
In particular, simple detectors based on VFM features consistently achieve better generalization and improved robustness than more sophisticated detectors working on dedicated features \cite{uhlenbrock2025latent}.
Especially significant is the high robustness to blurring, resizing and JPEG compression, as well as to low-level fingerprint injection, which suggests that these features do not capture (only) traditional high-frequency artifacts but deeper low-frequency cues \cite{cozzolino2024raising,guillaro2025bfree,huang2025rethinking,zhou2026simplicity}.
These cues have often been referred to as ``semantic'' in the literature, but their true nature remains poorly understood.
In this work, we shed light on this issue and investigate which image components are exploited by foundation-model-based detectors, revealing a distribution mismatch between real and generated images that extends to medium and low frequencies and is unrelated to image semantics.

\vspace{3mm}
\noindent
{\bf Diffusion-generated vs real images.}
Assessing whether generated images follow the distribution of real images is inherently difficult. The target distribution is accessible only through sampling, while the distribution learned by a generative model is often specified only implicitly, preventing exact likelihood evaluation. On the other hand, even likelihood-based criteria are only partially informative,
since high-likelihood samples are not typical in high-dimensional spaces \cite{cover2009elements}, that is, 
high-likelihood images are not perceptually pleasing or realistic \cite{theis2016note}. 
Diffusion models, in particular, suffer from known mismatch: the distribution reached at the end of the (finite) forward noising process does not exactly match the distribution used to start the reverse sampling, usually a white Gaussian noise. This prior mismatch causes the generated samples to deviate from the true data distribution, even when the score model is learned accurately \cite{wang2024solving}.
Practical evidence comes from studies on synthetic dataset: models trained on generated images underperform those trained on real data when evaluated on real test sets, suggesting that synthetic images do not fully capture real-world variability \cite{geng2024unmet,shumailov2024ai,adamkiewicz2026whenpretty}.
Furthermore, direct measurements on spectral distributions show that diffusion-generated images differ significantly from real images at high frequencies \cite{corvi2023intriguing,adamkiewicz2026whenpretty}.
The present work provides evidence that this spectral mismatch extends well beyond this, involving also medium-low frequencies.

\vspace{3mm}
\noindent
{\bf Diffusion-based image inversion.}
In the context of diffusion models, image inversion is a process that takes an image $x$ and derives a noise vector $z$ that, when used as a seed by a given diffusion model, allows it to regenerate $x$ exactly.
In practice, regeneration is never perfect for a variety of reasons, but image inversion can still be leveraged to perform important forensic tasks.
In particular, DIRE \cite{Wang2023dire} observes that diffusion-generated images are typically reconstructed better than real images by a pre-trained diffusion model and uses the reconstruction error as a decision statistic.
DIRE's good results suggest that real and generated images occupy different regions with respect to the dynamics learned by a diffusion model and that generated images may be closer to the manifold learned by the model than real images.
This basic idea has been further explored in subsequent works, through analysis of the latent space \cite{cazenavette2024fakeinversion}, considering intermediate reconstruction steps \cite{wu2026explainable,vasilcoiu2025latte}, adopting a single denoising step \cite{luo2024lare2,wang2025diffusionimplicitdetector}, forcing the detection network to predict the reconstruction error \cite{zhong2025beyond}.
Unlike all these works, we do not blindly use reconstruction error as decision statistics, but instead force the inversion to ensure semantic identity with the original image. Under this constraint, we study the appearance and structure of forensic artifacts and their impact on the decision of foundation model-based detectors.

\section{Motivation}

This section reviews the evidence supporting the use of current vision foundation models as excellent feature extractors for AI-generated image detection.
We consider multiple pretrained models: 
OpenCLIP~\cite{ilharco2021openclip,radford2021learning}, MetaCLIP~1~\cite{xu2024demystifying}, 
MetaCLIP~2~\cite{chuang2025meta}, 
DINOv3~\cite{simeoni2025dinov3}, as feature extractors.
Only for DINOv3 we also explore different backbones together with a version based on LoRA~\cite{hu2022lora}.
We demonstrate that detectors based on these features deliver state-of-the-art performance.
Furthermore, even when trained on images from a single diffusion model, they generalize well to images generated by all other unseen diffusion models.
Finally, they prove extremely robust to all common types of image degradation, including distortions that strongly affect high-frequency content.

\begin{table}[b!]
    \caption{AUC performance of VFM-based detectors.}
    \centering
	\small
	\scalebox{0.62}{
    \setlength{\tabcolsep}{4pt}
    \begin{tabular}{cl|C{12mm}C{12mm}C{12mm}C{12mm}C{12mm}C{12mm}C{12mm}C{12mm}C{12mm}C{12mm}|C{12mm}}
      \toprule
 \multicolumn{2}{c|}{AUC} & SD 1.4 & SD 2.1 & SDXL & SD 3 & Flux & DALL·E 3 & Firefly & Midj.  & Scale-RAE & Pixel DiT & AVG \\
  \midrule
& OpenCLIP   & \g{89.6} & 76.2 & 72.7 & 87.9 & 84.7 & 95.0 & 89.5 & 77.7 & 91.6 & 83.6 & 84.9  \\
& MetaCLIP 1 & \g{97.4} & 88.8 & 95.1 & 82.2 & 94.6 & 97.3 & 85.1 & 93.7 & 97.0 & 95.1 & 92.6  \\
& MetaCLIP 2 & \g{98.9} & 97.5 & 99.1 & 93.6 & 96.7 & 96.9 & 92.9 & 96.5 & 99.5 & 96.8 & 96.9  \\
\midrule
\multirow{6}{*}{\rotatebox[origin=c]{90}{DINOv3}}
& ConvNeXt B & \g{90.2} & 86.2 & 95.3 & 76.0 & 92.2 & 98.0 & 64.2 & 90.7 & 94.8 & 94.5 & 88.2 \\
& ConvNeXt L & \g{93.5} & 92.7 & 97.7 & 80.5 & 94.5 & 98.5 & 68.5 & 93.6 & 98.9 & 97.1 & 91.5 \\
& ViT-B/16   & \g{86.0} & 81.6 & 93.0 & 79.6 & 96.4 & 98.8 & 73.2 & 93.4 & 96.3 & 97.3 & 89.6 \\
& ViT-L/16   & \g{97.9} & 97.3 & 98.7 & 81.1 & 96.1 & 99.5 & 76.9 & 94.6 & 99.9 & 98.9 & 94.1 \\
& ViT-7B/16  & \g{100.} & 99.9 & 100. & 96.2 & 99.6 & 100. & 92.2 & 99.7 & 100. & 100. & 98.8 \\
& + LoRA     & \g{100.} & 100. & 100. & 98.7 & 99.2 & 99.9 & 97.4 & 99.7 & 100. & 100. & 99.5 \\
 \bottomrule
    \end{tabular}
}
    \label{tab:generalization}
\end{table}

\vspace{3mm}
\noindent
{\bf Generalization analysis.}
For all models, 
input images are resized to $224\times 224$ pixels, with no data augmentation, 
backbone parameters are frozen, 
and a linear head is optimized for binary classification 
(training at $2$ epochs, AdamW optimizer, learning rate $10^{-3}$, batch size $128$).
The training set is taken from the unbiased GenImage dataset \cite{grommelt2024fake}, 
with 162K fake images generated using Stable Diffusion 1.4 \cite{stablediffusion}, and as many real images taken from the ImageNet dataset \cite{deng2009imagenet}.
For testing, real images are taken from a different dataset, the RAISE dataset \cite{nguyen2015raise}, to reduce the risk of dataset-specific bias.
The analysis covers ten generators: 
SD 1.4, SD 2~\cite{stablediffusion}, 
SDXL~\cite{podell2024sdxl}, 
SD3~\cite{stablediffusion3}, 
Flux~\cite{flux1},
DALL-E 3~\cite{dalle3}, 
Firefly~\cite{firefly}, 
Midjourney~\cite{midjourney}, 
Scale-RAE~\cite{tong2026scaling}, and 
PixelDiT~\cite{yu2026pixeldit}. 
Such synthetic images are taken from \cite{bammey2023synthbuster,guillaro2025bfree}, 
except for Scale-RAE, and PixelDiT images which we generated ourselves.

Table \ref{tab:generalization} 
presents all results in terms of Area Under the ROC Curve (AUC). Only the first (gray) column shows the results for a model seen in training, while all the others refer to out-of-training models and demonstrate generalization ability. Overall, performance improves steadily along the rows, that is, with model size and backbone capacity.
Smaller models, such as OpenCLIP and MetaCLIP, already perform well, but generalize poorly on some DMs (SDXL, Flux, Firefly, and PixelDiT). In contrast, larger models show much higher transferability. Among the DINOv3 variants, performance increases steadily from ConvNeXt-B to ViT-7B/16, with the latter achieving an average AUC close to 99. LoRA achieves the best overall generalization, with an average AUC of 99.5 and near-perfect performance on most generators.

It is worth underlining that Scale-RAE and PixelDiT adopt substantially different architectures than Stable Diffusion 1.4. Scale-RAE replaces the traditional latent-space VAE with a Representation Autoencoder based on pre-trained visual representations, while PixelDiT removes the autoencoder altogether and performs diffusion directly in the pixel space using a two-level transformer architecture. Furthermore, these models are very recent: PixelDiT was released in November 2025 and Scale-RAE in January 2026. This means that the excellent performance observed on these generators cannot be explained by direct or indirect data leakage from the detector's training set. Rather, they suggest that the generated images share common features that foundation models are able to capture and exploit even when the detector is trained only on SD 1.4 images.

\begin{figure}[t!]
    \centering
    \includegraphics[width=1.0\linewidth,clip,trim=0 0 0 0]{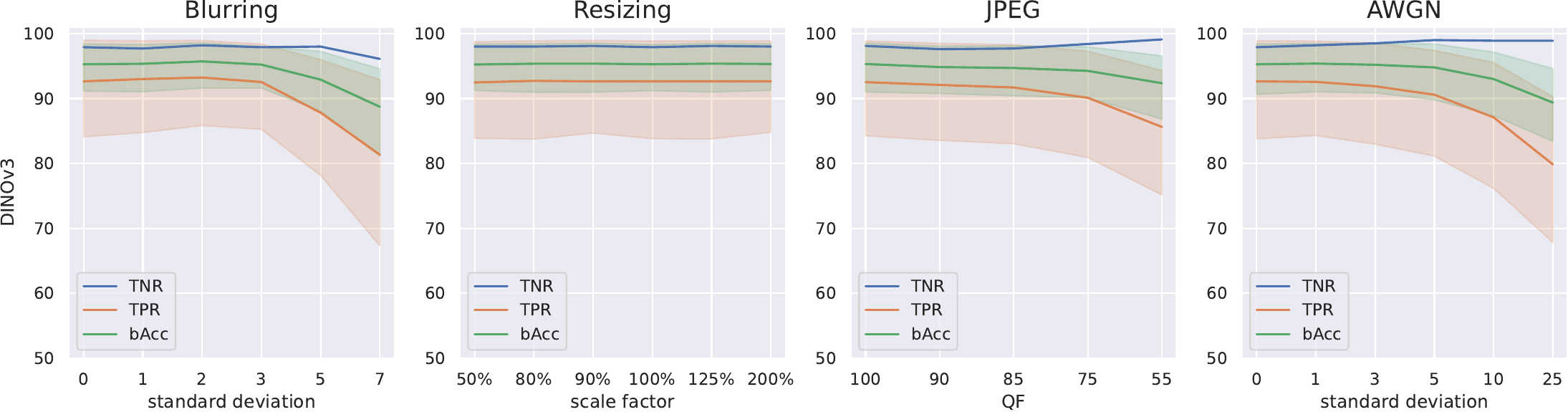}
    \caption{Performance of a DINOv3-based detector under different post-processing operations. Results are in terms of true positive rate (TPR), true negative rate (TNR), and balanced Accuracy (bAcc). Shaded bands indicate 95\% confidence intervals (CIs). The detector is robust to all degradations over a wide range of parameter values.}
    \label{fig:robust}
\end{figure}

\vspace{3mm}
\noindent
{\bf Robustness analysis.}
We now turn to analyzing robustness to image distortions, with a focus on degradations that remove or severely attenuate high-frequency content. This is particularly relevant in digital forensics, as images shared online are commonly resized, compressed, or otherwise processed by social media platforms. Conventional detectors often experience significant performance degradation following these operations, suggesting they rely heavily on subtle forensic traces that can be easily weakened or removed by standard post-processing \cite{tariang2024synthetic}.

Figure~\ref{fig:robust} shows the performance of the detector based on the DINOv3 (Vit-7B/16 backbone) features on the same test set used in the previous paragraph, under four common image degradations: blurring, resizing, JPEG compression, and white Gaussian noise addition. Overall, the detector is remarkably stable under moderate perturbations, with TNR, TPR, and balanced accuracy remaining close to the original values in most settings. For JPEG compression and resizing, the balanced accuracy remains relatively stable even under severe distortions (e.g., JPEG quality factor 55). Significant impairments are observed only under strong blurring and noising, when the TPR decreases noticeably. A possible interpretation is that such severe degradations corrupt the features that most qualify the image as synthetic. In general, these results support the idea that the detector does not rely only on mid- and high-frequency artifacts. If it did, performance would worsen under compression or resizing. Instead, the relatively stable behavior indicates that the model captures more robust cues that are degraded only when distortions become severe enough to alter broader image structures or suppress informative spectral components. Remarkably, the detector achieves an almost negligible false alarm rate in all situations.

It is also worth noting that this behavior makes the detector robust to biases introduced by image coding or post-processing pipelines. Because these distortions primarily affect high-frequency components, they have limited influence on decisions that appear to be based on deeper discrepancies between real and synthetic images, as shown in \cite{zhou2026simplicity}.
For the same reason, the detector is largely insensitive to traces left by the auto-encoder alone \cite{cozzolino2024raising}, since these also reside primarily in the high-frequency portion of the image spectrum.
Indeed a laundering operation using the VAE of SD 1.4 reduces the true negative rate of DINOv3-based detector from 97.9\% to 95.5\%. 

\section{What Do Foundation Models See?}

In the previous sections we argued that foundation models are able to see discriminative traces that reveal the synthetic nature of images generated by diffusion models, and that such traces reside in the low- to mid-frequency range of the spectrum.
However, just because of their low-frequency nature, such traces are not easily visualized.
At low frequencies, the image spectrum is dominated by the semantic content, 
and conventional filtering techniques are unable to separate these two components, image and artifacts,
nor can they extract one component without disrupting the other.
To overcome this limitation and gain insights into what cues the foundation models see, we design a custom strategy based on the DDIM inversion procedure \cite{song2020denoising}.
Given a real test image, we use a diffusion model to generate a stream of images that all share the same semantic content but exhibit appreciable differences.
Then, we extract various statistics from these differences and study how they relate to the detector score, gaining eventually further evidence in support of our initial claim.
In the remainder of this section, we first recall the basic concepts of the DDIM generation process and its inversion, and then present the proposed analysis.

\begin{figure}[t!]
    \centering
    \includegraphics[width=1.0\linewidth,page=2,clip,trim=0 110 0 0]{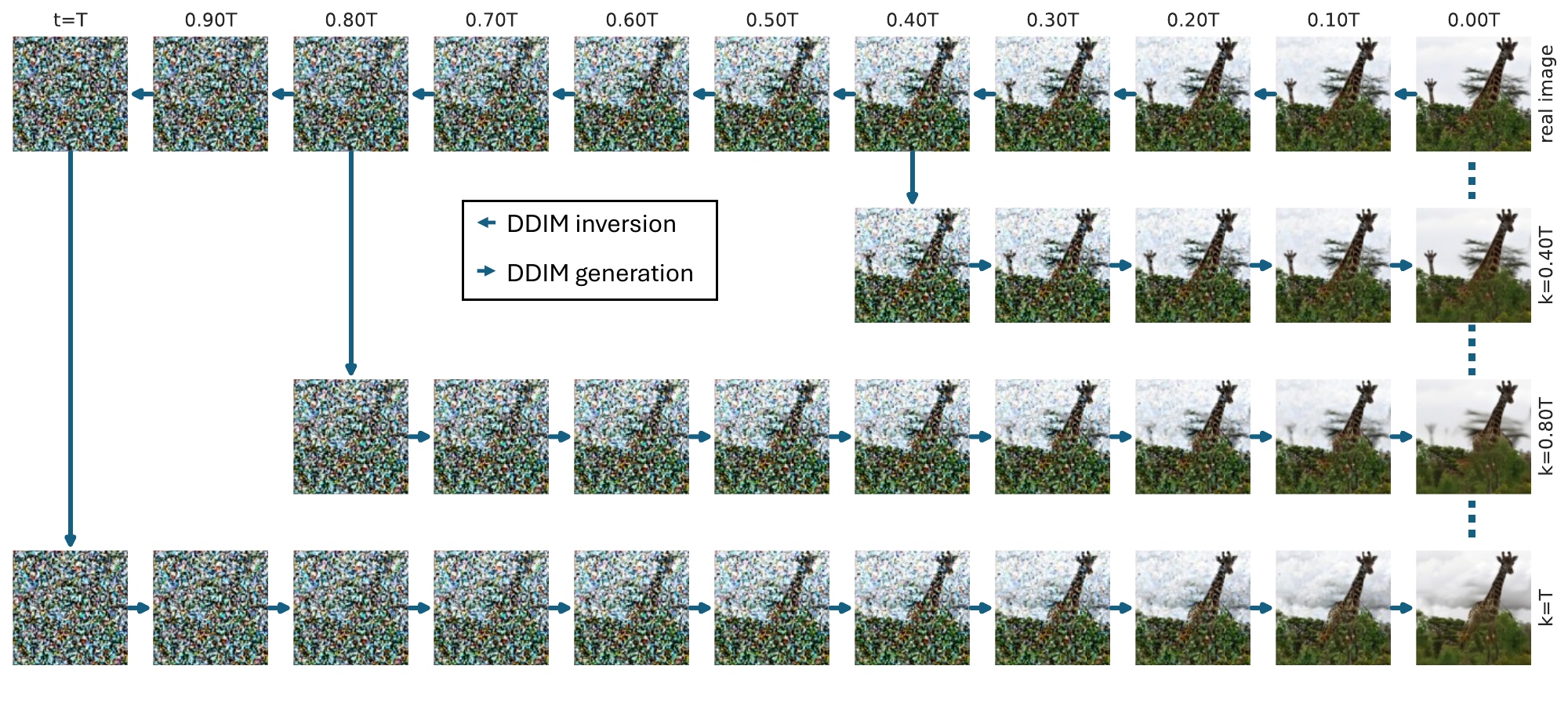}
    \caption{Diffusion process for an image sample. Starting from a real image (top-right), noise is progressively added in latent space using the DDIM inversion procedure \cite{song2020denoising}, obtaining noisy samples at various timesteps $t$ (first row), until the sample approaches an almost pure-noise distribution at $t=1.0T$. For each selected timestep $k$ (following rows) the reverse denoising process produces a different regenerated image.}
    \label{fig:schema}
\end{figure}

\subsection{Background}

Diffusion models learn to synthesize images by gradually reversing a forward noise-corruption process.
In the forward process, noise is progressively added to a clean image until it becomes approximately white Gaussian noise.
To invert this process, a denoiser is trained to predict the noise added at each timestep, 
thus allowing the generation of new samples starting from pure noise.

Denoising Diffusion Implicit Models (DDIMs) \cite{song2020denoising} 
transform noise into a clean image adopting a deterministic sampling strategy.
The DDIM generation process starts from a sufficiently large timestep $t=T$ (e.g., $T=1000$) and proceeds until $t=0$.
Each timestep is associated with a monotonic cumulative noise strength $\beta_t$, where $\beta_0=0$ and $\beta_T \approx 1$.
The initial sample $x_T$ is randomly drawn from a normal distribution. 
Then, at each step, a new sample $x_\tau$, with $\tau<t$, is generated as:
\begin{equation}
    \begin{split}
    x_\tau &= \sqrt{1-\beta_\tau} \cdot \left( \frac{x_t - \sqrt{\beta_t}
    \cdot\hat{\epsilon}_{\theta}(x_t, t)}{ \sqrt{1-\beta_t }} \right) +  \sqrt{\beta_\tau} \cdot \hat{\epsilon}_{\theta}(x_t, t) \\
    \end{split}
    \label{equ:DDIM}
\end{equation}
where $\hat{\epsilon}_{\theta}(x_t, t)$ is the output of a pretrained denoising model, 
which predicts the noise added to sample $x_t$ at a known timestep $t$.
Then the new timestep $\tau$ becomes the current timestep $t$ and the iterations proceed until $t=0$, where $x_0$ represents the generated sample.

Compared to previous diffusion sampling strategies, 
DDIMs allow image generation with fewer denoising steps while preserving high sample quality.
Its more relevant peculiarity, however, is that the noise added at each step is not random but strictly deterministic.
As a consequence, with DDIM, the same initial noise sample always produces the same output image.
This property allows us to reverse this path and recover the initial noise sample from a given real image.
In \cite{song2020denoising} an approximate strategy was also proposed to this end, known as DDIM inversion.
In this case, $x_0$ is the initial sample and $x_T$ the target, but new samples $x_\tau$, with $\tau>t$, keep being generated using Eq.~(\ref{equ:DDIM}).

\begin{figure}[t!]
    \centering
    \includegraphics[width=1.0\linewidth]{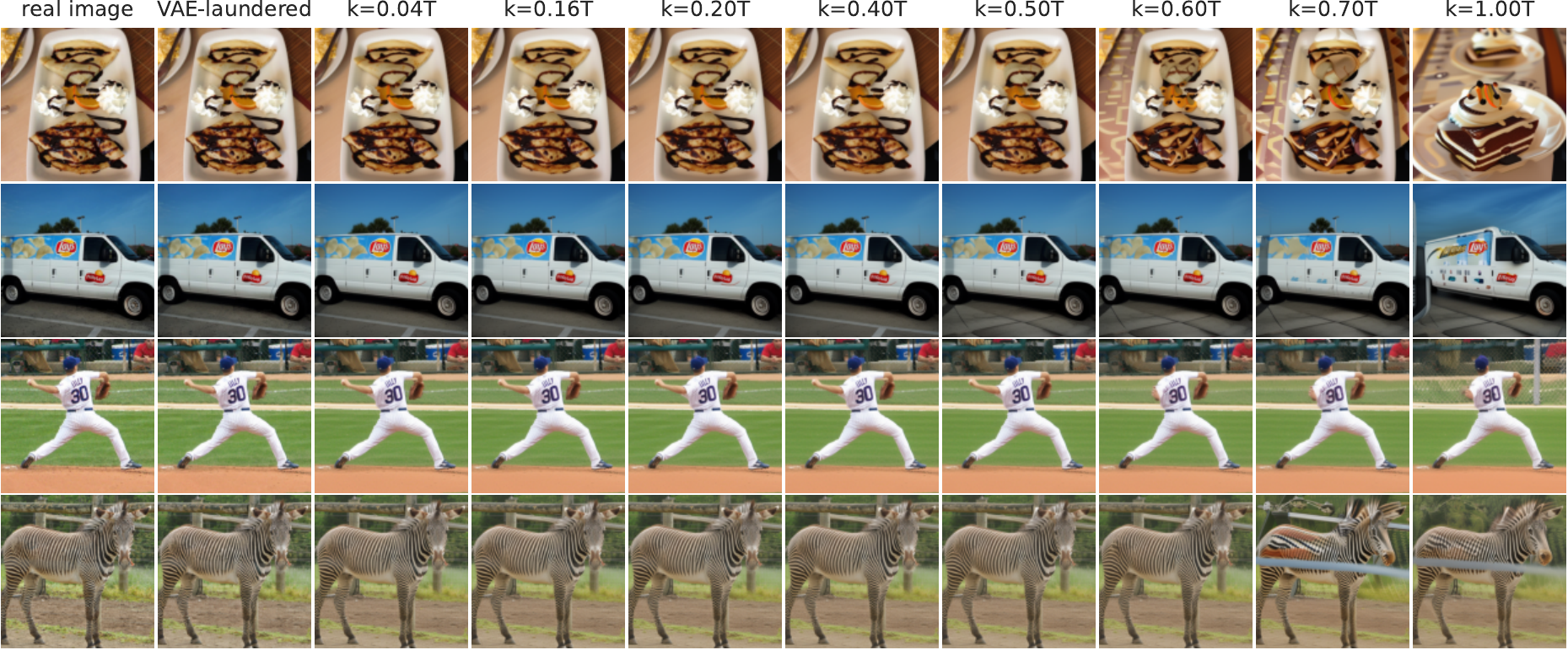}
    \caption{Examples of DDIM-based regeneration sequences. Starting from a real image, we apply DDIM inversion up to different timesteps \(k\), and then regenerate the image through the corresponding DDIM sampling process. For small and intermediate values of \(k\), the regenerated images preserve the semantic content of the original sample, while progressively introducing diffusion-related traces. For large values of \(k\), especially close to \(T\), stronger deviations from the original content may appear.}
    \label{fig:examples}
\end{figure}

\subsection{Methodology}

Our objective, here, is to generate a sequence of DM images
all semantically identical to a reference real image but all different from it at pixel level,
so as to study which non-semantic artifacts drive the decisions of a VFM-based detector.
For each real image,
we generate multiple reconstructed versions
by applying DDIM inversion at various depths followed by the corresponding generation procedure.
With reference to Figure~\ref{fig:schema},
the first row shows the DDIM inversion process where noise is gradually added to the reference image\footnote{Note that all processes described here take place in the latent space, in the figure we show the corresponding images obtained after decoding.} (top-right) until at the final timestep, $k=T$, the noisiest sample is obtained (top-left).
Now, starting from this sample we can use the standard DDIM generation process to produce a clean image (bottom row) which should be identical to the original image.
However, the DDIM inversion process is only approximate and starting from this very noisy sample the result is very often different from the original, not only at pixel level but also semantically.
So, to overcome this problem, we select closer starting points, $k<T$, to run DDIM generation (intermediate rows in the figure), ending up with images more similar to the original.
These generated images are shown in the rightmost column of the figure.
Some of them, typically, those with remote starting points $k \simeq T$, will differ semantically from the reference image, but the majority, with closer starting points, $k \ll T$, will exhibit the same semantic content.

\begin{figure}[t!]
    \centering
    \includegraphics[width=1.0\linewidth]{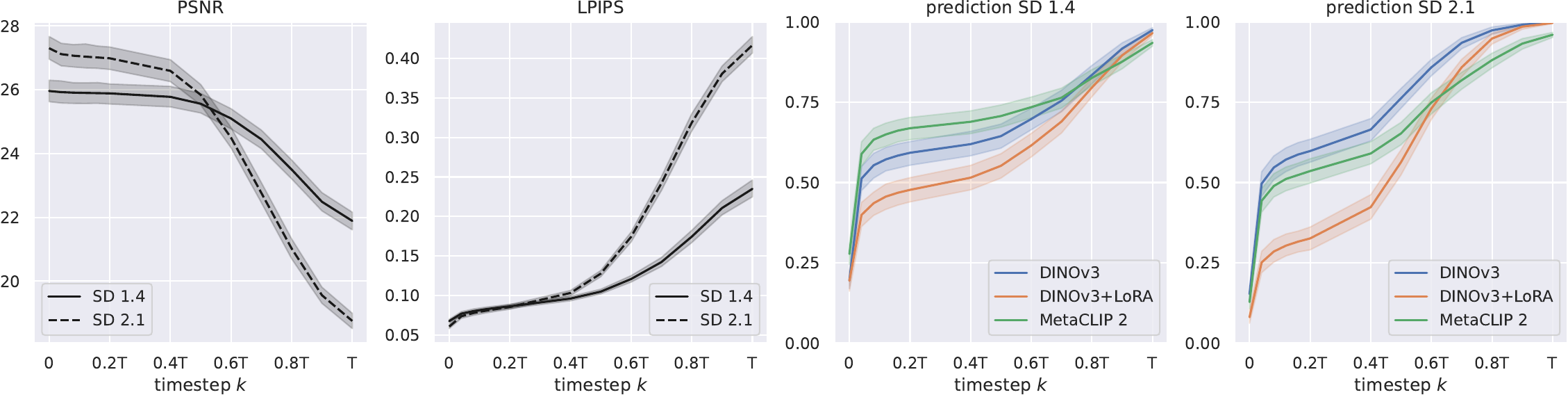}
    \caption{Effect of the inversion timestep $k$ on image reconstruction and detector response.
    Left: PSNR and LPIPS between the original real images and their DDIM-regenerated versions for SD~1.4 and SD~2.1. Both metrics remain relatively stable for small and intermediate values of $k$, then degrade as $k$ approaches $T$, indicating stronger deviations from the original image. Right: average detector prediction for images regenerated with SD~1.4 and SD~2.1. The detector score increases sharply with $k$ well before the onset of appreciable objective degradations (PSNR, LPIPS), suggesting that diffusion-related traces are introduced before semantic changes become visible.
    Shaded bands indicate 95\% confidence intervals (CIs).
    }
    \label{fig:prediction}
\end{figure}

In Figure~\ref{fig:examples} we show some examples of the generated sequences.
For low values of $k$, the regenerated images are almost identical to the reference.
In contrast, when $k=T$, the regenerated image differs significantly from it, while still preserving its high-level (textual) semantics and overall visual style.
Intermediate values of $k$ provide a sequence of regenerated variants of the same real image, with progressively increasing traces of their synthetic origin.
To study systematically the effects of such traces, we experiment on 1,000 pristine images of size $512 \times 512$ pixels extracted from the MS-COCO dataset~\cite{lin2014mscoco}.
We consider two diffusion models, Stable Diffusion 1.4 (SD 1.4) and Stable Diffusion 2.1 (SD 2.1). In both cases, the maximum number of steps is 50, and the process is text-conditioned using the caption associated with the pristine image, with guidance scale of 1.
Higher guidance values tend to overly degrade the inversion result, reducing its semantic consistency with the reference real image \cite{han2024proxedit,zeng2026does}.

In Figure~\ref{fig:prediction} (left), we report PSNR and LPIPS as a function of the inversion timestep $k$.
They remain almost stable for $k<0.5T$ and then exhibit a decreasing trend.
This confirms that, for low values of $k$, the regenerated images are nearly identical to the original image.
In Figure~\ref{fig:prediction} (right),
we also show that the detector response  increases steadily, and often very fast, as $k$ grows, with prediction exceeding $0.5$ (synthetic) even for low values of $k$.
Since regenerated images are not part of the training data of detector, the observed response is unlikely to be driven by artifacts specific to the inversion procedure. Instead, it suggests that the detector is responding to traces introduced by the generative process itself.
Moreover, the detector does not seem to rely on style and semantics to make its decision, considering that for low values of $k$ the regenerated images closely resemble the original ones under these aspects.
In the following, we carry out additional analyses to understand what makes synthetic images distinguishable from real ones.

\begin{figure}[t!]
    \centering
    \includegraphics[width=1.0\linewidth,page=3,clip,trim=0 250 0 0]{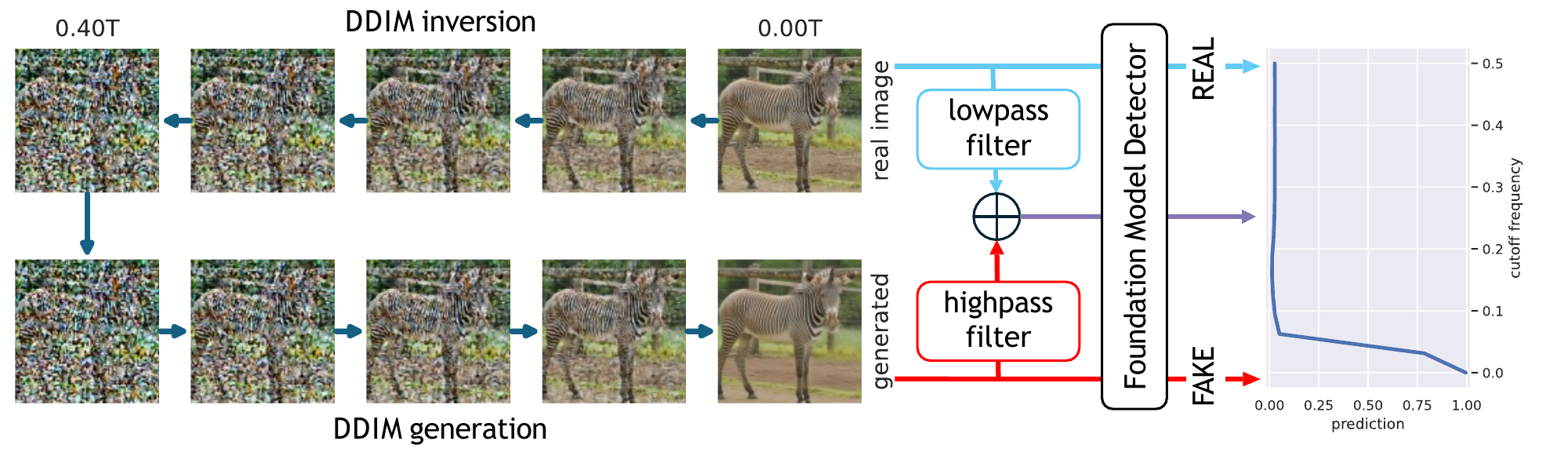}
    \caption{Overview of the frequency domain analysis.
Starting from a real image and its regenerated version at k = 0.4T with SD 2.1. Hybrid images are constructed by combining the low-frequency components of real image with the high-frequency components of the genereted one. On the right, the response of the VFM-based detector as a function of the cutoff frequency shows that the detector relies mainly on low-frequency image cues to make the decision.}
    \label{fig:schema_freq}
\end{figure}

\subsection{Analysis in the frequency domain}

In this section, we aim to identify the frequency components that are most relevant for the classification decision.
To this end, we consider each real image together with its near-identical corresponding regenerated version at $k=0.4T$ with SD 2.1.
We then construct hybrid images in the frequency domain by exchanging low- and high-frequency components between the real and regenerated samples, see Figure \ref{fig:schema_freq}.
Note that the high similarity between the two source images reduces the risk that the exchange process introduces artifacts into the hybrid image.
Specifically, we combine the low-frequency content of the real image with the high-frequency content of the regenerated one, and vice versa.
The resulting images are then evaluated by the VFM-based detectors to assess which frequency bands carry the most discriminative cues.
This analysis is restricted to those samples for which the regenerated image is classified as fake more than 50\% of the total.

\begin{figure}[t!]
    \centering
    \includegraphics[width=1.0\linewidth]{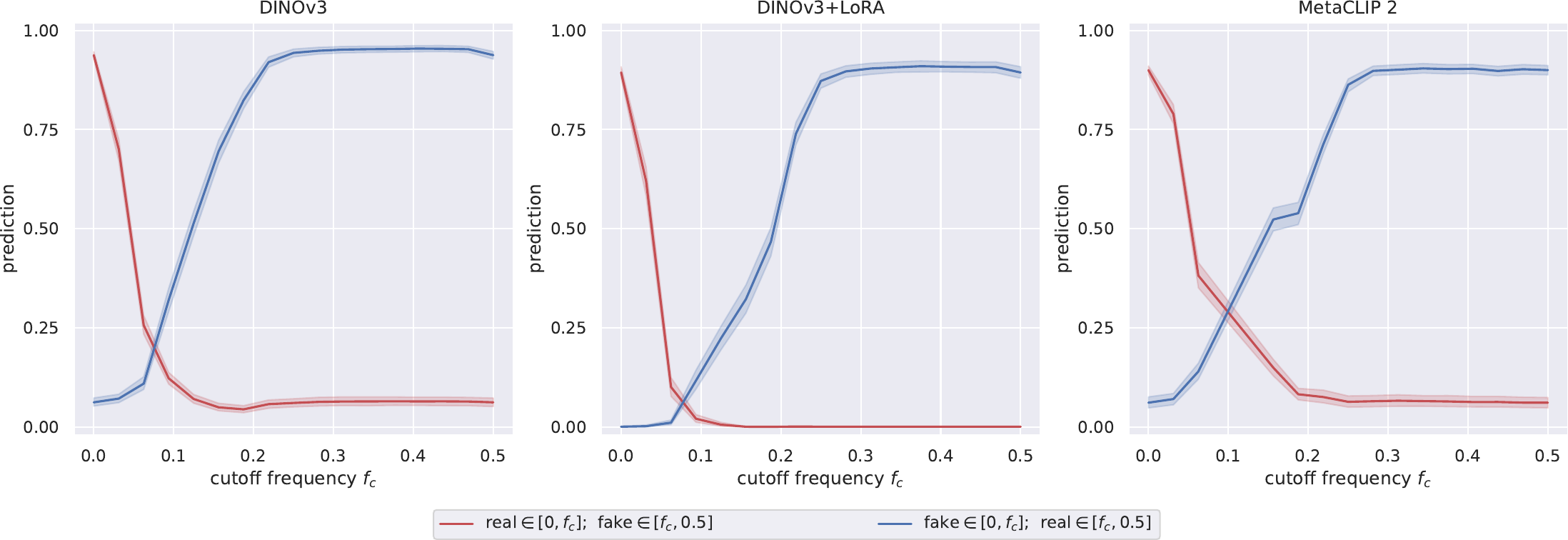}
    \caption{Frequency-based analysis of the detectors response. We build hybrid images by exchanging low- and high-frequency components between real images and their regenerated counterparts at $k=0.4T$ with SD 2.1. The red curves correspond to images with real low frequencies and regenerated high frequencies, while the blue curves correspond to regenerated low frequencies and real high frequencies. 
    }
    \label{fig:freq_analysis}
\end{figure}

Figure~\ref{fig:freq_analysis} reports the detector predictions obtained on the hybrid images as a function of the cutoff frequency for DINOv3, DINOv3+LoRA and MetaCLIP2. 
The red curve corresponds to a fake image ($f_c = 0$) that is progressively modified so as to include low-frequency components of the real image extracted with a filter with cut-off frequency equal to $f_c$, while the blue curve corresponds to the inverse combination.
Across all considered detectors, the response is mainly driven by the low-frequency content.
When the low-frequency components are taken from the generated image and the high-frequency components from the real image, the prediction rapidly increases as the cutoff frequency grows, up to the frequency of 0.25 where it remains constant.
Likewise, when the low-frequency components are taken from the real image and only the high-frequency components from the generated one, the detector output tends to decrease, indicating a classification in the real class.
These results suggest that the synthetic traces exploited by the these detectors are primarily localized in the lower-frequencies of the image. It is particularly interesting that injecting frequencies within the limited range $([0, 0.05])$ shifts the model’s decision from fake to real. 
This indicates that the generation process produces low-frequency components that are not fully coherent with the natural image manifold. 
This is also consistent with \cite{guillaro2025bfree}, where training data are built from real images and their self-conditioned counterparts. Since these pairs differ at the lowest frequencies, detectors can exploit inconsistencies across a broad frequency range rather than relying only on high-frequency traces.

\subsection{Analysis in the latent space}

\begin{figure}[t!]
    \centering
    \includegraphics[width=1.0\linewidth]{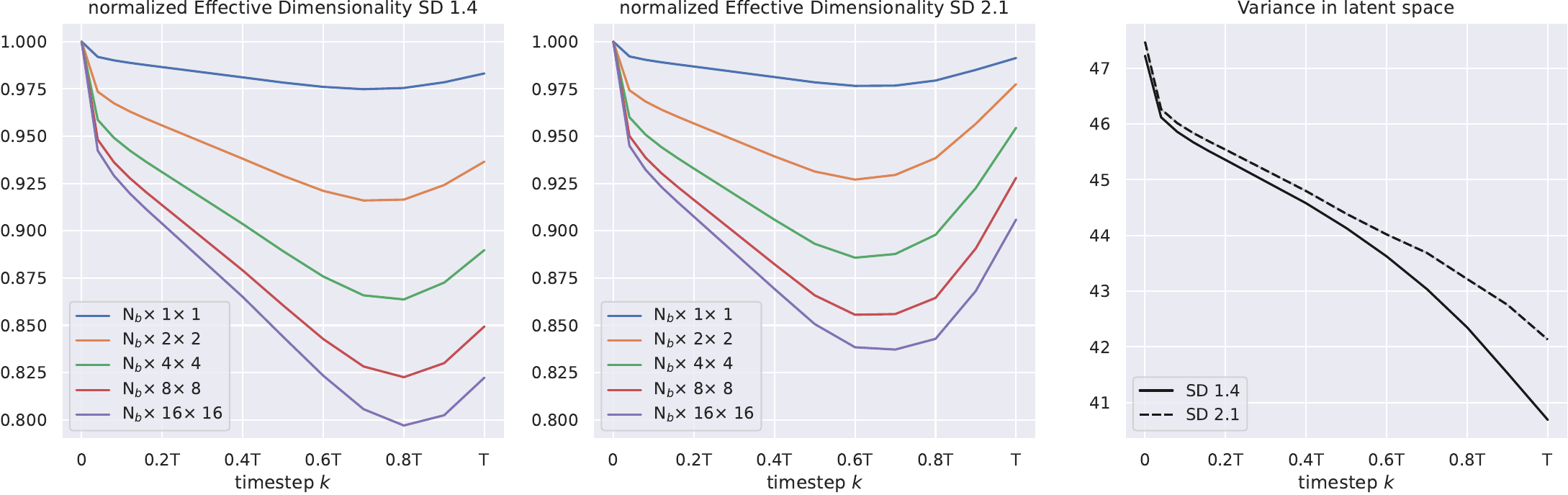}
    \caption{Latent-space variability analysis as a function of the inversion timestep $k$. Left and Middle: normalized effective dimensionality for SD~1.4 and SD~2.1, respectively, computed on latent blocks of different spatial sizes. In both models, the effective dimensionality decreases with $k$, indicating that the regenerated samples span a lower-dimensional latent distribution than the original real images. Right: average latent variance for SD~1.4 and SD~2.1, showing a progressive reduction as $k$ increases. }
    \label{fig:latent}
\end{figure}

Here we evaluate several measures in latent-space.
We start from the re-generated latent samples $z^i_k \in \mathbb{R}^{N_b \times N_r \times N_c}$, where $i$ is the image index, $k$ is the inversion timestep, and $N_b$, $N_r$ and $N_c$ are the number of channels, rows and columns of latent vector.
We compute the effective dimensionality \cite{roy2007effective} of the latent representations.
This measure estimates the intrinsic dimensionality of data and is therefore useful to quantify their variability.
In detail, we evaluate the effective dimensionality on latent-space blocks of size $N_b \times s \times s$, with $s \in \{1,2,4,8,16\}$. 
For each block-size $s$ and inversion timestep $k$, we consider all blocks of our dataset and compute the associated covariance matrix $\mathbf{C}_{s,k} \in \mathbb{R}^{D \times D}$, with dimensionality $D=s^2N_b$.
Let $\lambda_1,\lambda_2,\ldots,\lambda_D$ be the eigenvalues of $\mathbf{C}_{s,k}$, the effective dimensionality (ED), proposed in \cite{roy2007effective}, is then defined as
\begin{equation}
\mathrm{ED} = \exp\left(-\sum_{j=1}^{D} p_j \ln   p_j \right) \;\;\; \mathrm{with} \;\; p_j = \lambda_j / \sum_{i=1}^{D} \lambda_i
\end{equation}
If the variance is uniformly distributed across all directions, the effective dimensionality equals $D$. Conversely, if most of the variance is concentrated along a few dominant directions, the ED is much lower than $D$.
Figure~\ref{fig:latent} (left-mid) reports the effective dimensionality as a function of the inversion timestep $k$ for both SD~1.4 and SD~2.1. Specifically, values are normalized to the value for $k=0$, to allow comparison across different block sizes $s$. For both models, the normalized value decreases as $k$ increases, reaching a minimum around $k \simeq 0.75$. It then partially recovers, but never returns to its initial value at $k=0$. This increase at large $k$, where the denoising process starts from highly noisy latent states, is likely due to stronger alterations of the original image content during generation.

To further prove this effect, we consider the averaged variance of blocks of dimension $N_b\times 4\times 4$ (other dimensions follow the same trend). 
In Figure~\ref{fig:latent} (right), we can observe that variance decreases as $k$ increases, implying a lower variability of the generated data.
Overall, these results suggest that the generation process does not fully reproduce the variability present in real data, but concentrates content in a reduced number of dominant directions.
This observation is consistent with recent studies \cite{shumailov2024ai,dombrowski2025image} showing that generative models often cover only a subset of the real data distribution, producing samples with lower diversity and missing parts of the distributional tail.

\section{Discussion}

\vspace{3mm}
\noindent
{\bf Conclusions.}
Our analyses provide evidence that the strong performance of foundation-model-based detectors is only marginally related to semantic visual failures or high-frequency artifacts. Instead, it mostly relies on traces at the mid-low frequency range of the spectrum originated by the DM generation process. By exploiting DDIM inversion, we generate image sequences that preserve the content of the original real image while progressively introducing the effect of the diffusion generation process. The detector response increases with the inversion timestep even when the generated images are semantically close to the original ones, suggesting that the decision is not driven by semantic traces but rather by high-level scene inconsistencies.

The frequency-swapping analysis shows that detector responses are mainly driven by the low-frequency components of regenerated images. When the low frequencies come from the generated image, the sample is still classified as fake; when they come from the real image, the prediction shifts toward real. This suggests that foundation models exploit distributional discrepancies between real and diffusion-generated images that extend beyond high frequencies to the low- and mid-frequency ranges.
The latent-space analysis provides complementary evidence. As the inversion timestep increases, both average latent variance and effective dimensionality decrease, indicating that regeneration does not fully preserve the variability of real data but uses a smaller number of dominant directions. This supports the view that diffusion-generated images occupy a restricted region of the natural-image distribution.
These findings help explain why detectors based on large vision foundation models generalize well across generators and remain robust to common image degradations. 
They are mostly unaffected by resizing, compression, blurring, or laundering operations simply because their decisions are not driven by subtle high-frequency artifacts.

\vspace{3mm}
\noindent
{\bf Future work.}
This work focuses on diffusion-generated images and on detectors built from frozen vision foundation models followed by a simple classifier. Although this setting allows us to isolate the role of pretrained representations, it is limited to two variants of Stable Diffusion generators. Future work will extend the analysis to older GAN-based generators and to more recent diffusion-based generation pipelines, including those that apply the diffusion process directly in the pixel space \cite{yu2026pixeldit,chen2026l2p}. Although experiments show (see Table 1) that these latter generators are successfully detected as well, suggesting the existence of similar statistical signatures, it is important to understand whether the same or different cues are exploited. 
Furthermore, our results suggest the existence of common low-to-mid frequency statistical features shared by diffusion-generated images, but they do not yet provide a complete mathematical characterization of such traces. Understanding their origin in the diffusion process is an open problem. Answering this question could improve both the interpretability of foundation-model-based detectors and the design of more robust forensic methods for future generations of synthetic images.

\bibliographystyle{splncs04}
\bibliography{main}

@String(CVPR  = {CVPR})

@String(ICCV  = {ICCV})

@String(ECCV  = {ECCV})

@String(ECCVW  = {ECCV Workshops})

@String(NeurIPS = {NeurIPS})

@String(ICML  = {ICML})

@String(ICLR  = {ICLR})

@String(CVPRW = {CVPR Workshops})

@String(AAAI  = {AAAI})

@string(WIFS = {WIFS})

@String(EUSIPCO = {EUSIPCO})

@string(WACV = {WACV})

@inproceedings{ojha2023towards,
  title={Towards universal fake image detectors that generalize across generative models},
  author={Ojha, Utkarsh and Li, Yuheng and Lee, Yong Jae},
  booktitle=cvpr,
  pages={24480--24489},
  year={2023}
}

@inproceedings{cozzolino2024raising,
  title={{Raising the Bar of AI-generated Image Detection with CLIP}},
  author={Cozzolino, Davide and Poggi, Giovanni and Corvi, Riccardo and Nie{\ss}ner, Matthias and Verdoliva, Luisa},
  booktitle=cvprw,
  pages={4356--4366},
  year={2024}
}

@inproceedings{guillaro2025bfree,
  title={{A Bias-Free Training Paradigm for More General AI-generated Image Detection}},
  author={Fabrizio Guillaro and Giada Zingarini and Ben Usman and Avneesh Sud and Davide Cozzolino and Luisa Verdoliva},
  booktitle=cvpr,
  year={2025}
}

@inproceedings{sha2022fake,
author = {Sha, Zeyang and Li, Zheng and Yu, Ning and Zhang, Yang},
title={{DE-FAKE: Detection and Attribution of Fake Images Generated by Text-to-Image Diffusion Models}},
year = {2023},
booktitle = {ACM SIGSAC Conference on Computer and Communications Security},
pages = {3418–3432},
numpages = {15},
}

@inproceedings{corvi2025seeing,
  title={{Seeing What Matters: Generalizable AI-generated Video Detection with Forensic-Oriented Augmentation}},
  author={Riccardo Corvi and Davide Cozzolino and Ekta Prashnani and Shalini De Mello and Koki Nagano and Luisa Verdoliva},
  booktitle=neurips,
  year={2025}
}

@inproceedings{wang2023dire,
  title={{DIRE for diffusion-generated image detection}},
  author={Wang, Zhendong and Bao, Jianmin and Zhou, Wengang and Wang, Weilun and Hu, Hezhen and Chen, Hong and Li, Houqiang},
  booktitle=iccv,
  pages={22445--22455},
  year={2023}
}

@inproceedings{dombrowski2025image,
  title     = {Image Generation Diversity Issues and How to Tame Them},
  author    = {Mischa Dombrowski and Weitong Zhang and Sarah Cechnicka and Hadrien Reynaud and Bernhard Kainz},
  booktitle = cvpr,
  year      = {2025}
}

@article{shumailov2024ai,
  title   = {{AI models collapse when trained on recursively generated data}},
  author  = {Ilia Shumailov and Zakhar Shumaylov and Yiren Zhao and Nicolas Papernot and Ross Anderson and Yarin Gal},
  journal = {Nature},
  volume  = {631},
  pages   = {755--759},
  year    = {2024}
}

@inproceedings{song2020denoising,
  title={{Denoising Diffusion Implicit Models}},
  author={Song, Jiaming and Meng, Chenlin and Ermon, Stefano},
  year={2021},
  booktitle=iclr
}

@article{huang2025rethinking,
  title   = {{Rethinking Cross-Generator Image Forgery Detection through DINOv3}},
  author  = {Huang, Zhenglin and Li, Jason and Wen, Haiquan and Li, Tianxiao and Yang, Xi and Qi, Lu and Peng, Bei and Huang, Xiaowei and Yang, Ming-Hsuan and Cheng, Guangliang},
  journal = {arXiv preprint arXiv:2511.22471v1},
  year    = {2025}
}

@article{li2026spectral,
  title   = {{Spectral Tail Auxiliary Learning for AI-Generated Image Detection}},
  author  = {Li, Xingyi and Zhang, Jiahui and Li, Yiheng and Cao, Yun and Wang, Wenhao},
  journal = {arXiv preprint arXiv:2605.22751},
  year    = {2026}
}

@article{sun2026salartvqa,
  title   = {{SalArt-VQA: Diagnosing Whether VLMs Understand Salient Artifacts in Generated Images}},
  author  = {Sun, Xiaoxiao and Zhang, Ruotian and Huang, Junzhe and Burgess, James and Yeung-Levy, Serena},
  journal = {arXiv preprint arXiv:2606.12671},
  year    = {2026}
}

@inproceedings{uhlenbrock2025latent,
  title     = {{Latent Landscapes: Topology of the {CLIP} Feature Space for Synthetic Image Detection}},
  author    = {Uhlenbrock, Lea and Bergmann, Sandra and Riess, Christian},
  booktitle = EUSIPCO,
  year      = {2025}
}

@inproceedings{wang2025diffusionimplicitdetector,
  title     = {{Your Diffusion Model is an Implicit Synthetic Image Detector}},
  author    = {Wang, Xi and Kalogeiton, Vicky},
  booktitle = eccvw,
  volume    = {15643},
  pages     = {418--434},
  year      = {2025}
}

@inproceedings{zhong2025beyond,
  title     = {{Beyond Generation: A Diffusion-based Low-level Feature Extractor for Detecting AI-generated Images}},
  author    = {Zhong, Nan and Chen, Haoyu and Xu, Yiran and Qian, Zhenxing and Zhang, Xinpeng},
  booktitle = cvpr,
  pages     = {8258--8268},
  year      = {2025}
}

@inproceedings{wu2026explainable,
  title     = {{Explainable Synthetic Image Detection Through Diffusion Timestep Ensembling}},
  author    = {Wu, Yixin and Zhang, Feiran and Shi, Tianyuan and Yin, Ruicheng and Wang, Zhenghua and Gan, Zhenliang and Wang, Xiaohua and Lv, Changze and Zheng, Xiaoqing and Huang, Xuanjing},
  booktitle = aaai,
  volume    = {40},
  number    = {13},
  pages     = {10844--10852},
  year      = {2026}
}

@article{vasilcoiu2025latte,
  title   = {{LATTE: Latent Trajectory Embedding for Diffusion-Generated Image Detection}},
  author  = {Vasilcoiu, Ana and Najdenkoska, Ivona and Geradts, Zeno and Worring, Marcel},
  journal = {arXiv preprint arXiv:2507.03054v2},
  year    = {2025}
}

@article{wang2024solving,
  title   = {{Solving Prior Distribution Mismatch in Diffusion Models via Optimal Transport}},
  author  = {Zhanpeng Wang and Shenghao Li and Jiameng Che and Chen Wang and Shangling Jui and Na Lei and Zhongxuan Luo},
  journal = {arXiv preprint arXiv:2410.13431},
  year    = {2024}
}

@inproceedings{tan2026forendex,
  title     = {{ForenDeX: Unlocking Forensic Insights for Explainable AI-Generated Image Detection}},
  author    = {Tan, Chuangchuang and Wang, Jinglu and Ming, Xiang and Tao, Renshuai and Wei, Yunchao and Zhao, Yao and Lu, Yan},
  booktitle = cvpr,
  pages     = {6592--6601},
  year      = {2026}
}

@inproceedings{adamkiewicz2026whenpretty,
  title     = {{When Pretty Isn't Useful: Investigating Why Modern Text-to-Image Models Fail as Reliable Training Data Generators}},
  author    = {Adamkiewicz, Krzysztof and Moser, Brian B. and Frolov, Stanislav and Nauen, Tobias Christian and Raue, Federico and Dengel, Andreas},
  booktitle = cvpr,
  pages     = {36660--36669},
  year      = {2026}
}

@inproceedings{luo2024lare2,
  title     = {{LaRE}$^2$: Latent Reconstruction Error Based Method for Diffusion-Generated Image Detection},
  author    = {Luo, Yunpeng and Du, Junlong and Yan, Ke and Ding, Shouhong},
  booktitle = cvpr,
  pages     = {17006--17015},
  year      = {2024}
}

@inproceedings{geng2024unmet,
  title     = {The Unmet Promise of Synthetic Training Images: Using Retrieved Real Images Performs Better},
  author    = {Geng, Scott and Hsieh, Cheng-Yu and Ramanujan, Vivek and Wallingford, Matthew and Li, Chun-Liang and Koh, Pang Wei and Krishna, Ranjay},
  booktitle = NeurIPS,
  year      = {2024}
}

@InProceedings{cazenavette2024fakeinversion,
    author    = {Cazenavette, George and Sud, Avneesh and Leung, Thomas and Usman, Ben},
    title     = {{FakeInversion: Learning to Detect Images from Unseen Text-to-Image Models by Inverting Stable Diffusion}},
    booktitle = cvpr,
    month     = {June},
    year      = {2024},
    pages     = {10759-10769}
}

@article{zhou2026simplicity,
  title={{Simplicity Prevails: The Emergence of Generalizable AIGI Detection in Visual Foundation Models}},
  author={Yue Zhou and Xinan He and Kaiqing Lin and Bing Fan and Feng Ding and Bin Li},
  journal={arXiv preprint arXiv:2602.01738v2},
  year={2026}
}

@article{amerini2025deepfake,
  title   = {{Deepfake Media Forensics: Status and Future Challenges}},
  author  = {Amerini, Irene and Barni, Mauro and Battiato, Sebastiano and Bestagini, Paolo and Boato, Giulia and Bruni, Vittoria and Caldelli, Roberto and De Natale, Francesco and De Nicola, Rocco and Guarnera, Luca and Mandelli, Sara and Majid, Taiba and Marcialis, Gian Luca and Micheletto, Marco and Montibeller, Andrea and Orr{\`u}, Giulia and Ortis, Alessandro and Perazzo, Pericle and Puglisi, Giovanni and Purnekar, Nischay and Salvi, Davide and Tubaro, Stefano and Villari, Massimo and Vitulano, Domenico},
  journal = {Journal of Imaging},
  volume  = {11},
  number  = {3},
  pages   = {73},
  year    = {2025}
}

@inproceedings{li2024masksim,
  title     = {{MaskSim: Detection of Synthetic Images by Masked Spectrum Similarity Analysis}},
  author    = {Li, Yanhao and Bammey, Quentin and Gardella, Marina and Nikoukhah, Tina and Morel, Jean-Michel and Colom, Miguel and Grompone von Gioi, Rafael},
  booktitle = cvprw,
  pages     = {3855--3865},
  year      = {2024}
}

@inproceedings{theis2016note,
  title     = {A Note on the Evaluation of Generative Models},
  author    = {Theis, Lucas and van den Oord, A{\"a}ron and Bethge, Matthias},
  booktitle = iclr,
  year      = {2016}
}

@inproceedings{frank2020frequency,
  title={Leveraging Frequency Analysis for Deep Fake Image Recognition},
  author={Frank, Joel and Eisenhofer, Thorsten and Sch{\"o}nherr, Lea and Fischer, Asja and Kolossa, Dorothea and Holz, Thorsten},
  booktitle=icml,
  year={2020}
}

@inproceedings{durall2020upconv,
  title={{Watch Your Up-Convolution: CNN Based Generative Deep Neural Networks Are Failing to Reproduce Spectral Distributions}},
  author={Durall, Ricard and Keuper, Margret and Keuper, Janis},
  booktitle=cvpr,
  year={2020}
}

@inproceedings{karageorgiou2025spai,
  title={{Any-Resolution AI-Generated Image Detection by Spectral Learning}},
  author={Karageorgiou, Dimitrios and Papadopoulos, Symeon and Kompatsiaris, Ioannis and Gavves, Efstratios},
  booktitle=cvpr,
  year={2025}
}

@article{tariang2024synthetic,
  title={{Synthetic Image Verification in the Era of Generative AI: What Works and What Isn't There Yet}},
  author={Diangarti Tariang and Riccardo Corvi and Davide Cozzolino and Giovanni Poggi and Koki Nagano and Luisa Verdoliva},
  journal={IEEE Security \& Privacy},
  volume={22},
  pages={37--49},
  year={2024}
}

@article{bammey2023synthbuster,
  title={{Synthbuster: Towards detection of diffusion model generated images}},
  author={Bammey, Quentin},
  journal={IEEE Open Journal of Signal Processing},
  year={2023}
}

@article{bontcheva2024generative,
  title={{Generative AI and disinformation: recent advances, challenges, and opportunities}},
  author = {Bontcheva, Kalina and Papadopoulous, Symeon and Tsalakanidou, Filareti and Gallotti, Riccardo and Dutkiewicz, Lidia and Krack, No{\'e}mie and Teyssou, Denis and Nucci, Francesco Severio and Spangenberg, Jochen and Srba, Ivan and Aichroth, Patrick and Cuccovillo, Luca and Verdoliva, Luisa},
  year={2024}
}

@book{barrett2024identifying,
  title={{Identifying and Mitigating the Security Risks of Generative AI}},
  author = {Barrett, Clark and Boyd, Brad and Burzstein, Elie and Carlini, Nicholas and Chen, Brad and Choi, Jihye and Chowdhury, Amrita Roy and Christodorescu, Mihai and Datta, Anupam and Feizi, Soheil and Fisher, Kathleen and Hashimoto, Tatsunori and Hendrycks, Dan and Jha, Somesh and Kang, Daniel and Kerschbaum, Florian and Mitchell, Eric and Mitchell, John and Ramzan, Zulfikar and Shams, Khawaja and Song, Dawn and Taly, Ankur and Yang, Diyi},
 year={2024},
  publisher={Now Foundations and Trends}
}

@inproceedings{corvi2023intriguing,
  title={Intriguing properties of synthetic images: from generative adversarial networks to diffusion models},
  author={Corvi, Riccardo and Cozzolino, Davide and Poggi, Giovanni and Nagano, Koki and Verdoliva, Luisa},
  booktitle=cvprw,
  pages={973--982},
  year={2023}
}

@inproceedings{zhang2019detecting,
  title={{Detecting and Simulating Artifacts in GAN Fake Images}},
  author={Xu Zhang and Svebor Karaman and Shih-Fu Chang},
  booktitle=wifs,
  year={2019},
}

@INPROCEEDINGS{roy2007effective,
  author={Roy, Olivier and Vetterli, Martin},
  booktitle=EUSIPCO, 
  title={{The Effective Rank: a Measure of Effective Dimensionality}}, 
  year={2007},
  pages={606-610}
}

@inproceedings{radford2021learning,
  title={Learning Transferable Visual Models From Natural Language Supervision},
  author={Alec Radford and Jong Wook Kim and Chris Hallacy and Aditya Ramesh and Gabriel Goh and Sandhini Agarwal and Girish Sastry and Amanda Askell and Pamela Mishkin and  Jack Clark and Gretchen Krueger and Ilya Sutskever},
  booktitle=icml,
  year={2021}
}

@inproceedings{xu2024demystifying,
      title={{Demystifying CLIP Data}},
  author={Xu, Hu and Xie, Saining and Tan, Xiaoqing and Huang, Po-Yao and Howes, Russell and Sharma, Vasu and Li, Shang-Wen and Ghosh, Gargi and Zettlemoyer, Luke and Feichtenhofer, Christoph},
  booktitle=iclr,
  volume={2024},
  pages={47812--47831},
  year={2024}
}

@article{chuang2025meta,
  title={{MetaCLIP2: A Worldwide Scaling Recipe}},
  author    = {Chuang, Yung-Sung and Li, Yang and Wang, Dong and Yeh, Ching-Feng and Lyu, Kehan and Raghavendra, Ramya and Glass, Jim and Huang, Lifei and Weston, Jason E. and Zettlemoyer, Luke and Chen, Xinlei and Liu, Zhuang and Xie, Saining and Yih, Scott and Li, Shang-Wen and Xu, Hu},
 journal=NeurIPS,
  volume={38},
  pages={48009--48036},
  year={2025}
}

@article{simeoni2025dinov3,
  title={{DINOv3}},
  author={Sim{\'e}oni, Oriane and Vo, Huy V. and Seitzer, Maximilian and Baldassarre, Federico and Oquab, Maxime and Jose, Cijo and Khalidov, Vasil and Szafraniec, Marc and Yi, Seungeun and Ramamonjisoa, Micha{\"e}l and Massa, Francisco and Haziza, Daniel and Wehrstedt, Luca and Wang, Jianyuan and Darcet, Timoth{\'e}e and Moutakanni, Th{\'e}o and Sentana, Leonel and Roberts, Claire and Vedaldi, Andrea and Tolan, Jamie and Brandt, John and Couprie, Camille and Mairal, Julien and J{\'e}gou, Herv{\'e} and Labatut, Patrick and Bojanowski, Piotr},
  journal={arXiv preprint arXiv:2508.10104},
  year={2025}
}

@inproceedings{grommelt2024fake,
  title={{Fake or JPEG? Revealing Common Biases in Generated Image Detection Dataset}},
  author={Grommelt, Patrick and Weiss, Louis and Pfreundt, Franz-Josef and Keuper, Janis},
  booktitle=eccvw,
  pages={80--95},
  year={2024}
}

@inproceedings{hu2022lora,
  title={{LoRA: Low-Rank Adaptation of Large Language Models}},
  author={Edward J. Hu and Yelong Shen and Phillip Wallis and Zeyuan Allen-Zhu and Yuanzhi Li and Shean Wang and Lu Wang and Weizhu Chen},
  booktitle=iclr,
  year={2022},
}

@inproceedings{nguyen2015raise,
author = {Dang-Nguyen, Duc-Tien and Pasquini, Cecilia and Conotter, Valentina and Boato, Giulia},
title = {{RAISE: a raw images dataset for digital image forensics}},
year = {2015},
booktitle = {ACM Multimedia Systems Conference},
pages = {219–224}
}

@inproceedings{lin2014mscoco,
  title={{Microsoft COCO: Common Objects in Context}},
  author={Lin, Tsung-Yi and Maire, Michael and Belongie, Serge and Hays, James and Perona, Pietro and Ramanan, Deva and Doll{\'a}r, Piotr and Zitnick, C Lawrence},
  booktitle=eccv,
  pages={740--755},
  year={2014}
}

@misc{stablediffusion,
    author= {Rombach, R. and Blattmann, A. and Lorenz, D. and Esser, P. and Ommer, B.},
    title = "\url{https://github.com/CompVis/stable-diffusion}",
    journal={~},
    year={2022}
}

@inproceedings{podell2024sdxl,
  title={{SDXL: Improving Latent Diffusion Models for High-Resolution Image Synthesis}},
  author={Podell, Dustin and English, Zion and Lacey, Kyle and Blattmann, Andreas and Dockhorn, Tim and M{\"u}ller, Jonas and Penna, Joe and Rombach, Robin},
  booktitle=iclr,
  volume={2024},
  pages={1862--1874},
  year={2024}
}

@misc{dalle3,
    author= {OpenAI},
    title = "\url{https://openai.com/dall-e-3}",
    journal={~},
    year={2023}
}

@misc{midjourney,
    author= {Midjourney},
    title = "\url{https://www.midjourney.com/home}",
    journal={~},
    year={2023}
}

@misc{stablediffusion3,
    author= {{Stable Diffusion 3}},
    title = "\url{https://stability.ai/news/stable-diffusion-3}",
    journal={~},
    year={2024}
}

@misc{firefly,
    author= {Adobe Firefly},
    title = "\url{https://www.adobe.com/sensei/generative-ai/firefly.html}",
    journal={~},
    year={2023}
}

@misc{flux1,
    author={Black Forest Labs},
    year={2024},
    title="\url{https://github.com/black-forest-labs/flux}",
}

@misc{ilharco2021openclip,
  author= {Ilharco, Gabriel and Wortsman, Mitchell and Wightman, Ross and Gordon, Cade and Carlini, Nicholas and Taori, Rohan and Dave, Achal and others},
  title        = "OpenCLIP \url{https://doi.org/10.5281/zenodo.5143773}",
  month        = jul,
  year         = {2021},
}

@inproceedings{yu2026pixeldit,
  title={{PixelDiT: Pixel Diffusion Transformers for Image Generation}},
  author={Yu, Yongsheng and Xiong, Wei and Nie, Weili and Sheng, Yichen and Liu, Shiqiu and Luo, Jiebo},
  booktitle=cvpr,
  pages={14273--14282},
  year={2026}
}

@article{tong2026scaling,
  title={{Scaling Text-to-Image Diffusion Transformers with Representation Autoencoders}},
  author={Tong, Shengbang and Zheng, Boyang and Wang, Ziteng and Tang, Bingda and Ma, Nanye and Brown, Ellis and Yang, Jihan and Fergus, Rob and LeCun, Yann and Xie, Saining},
  journal={arXiv preprint arXiv:2601.16208},
  year={2026}
}

@ARTICLE{lin2024detecting,
  author={Li Lin and Neeraj Gupta and Yue Zhang and Hainan Ren and Chun-Hao Liu and Feng Ding and Xin Wang and Xin Li and Luisa Verdoliva and Shu Hu},
  journal={arXiv preprint arXiv:2402.00045},
  title={{Detecting multimedia generated by large AI models: A survey}}, 
  year={2024}
}

@article{chen2026l2p,
  title={{L2P: Unlocking Latent Potential for Pixel Generation}},
  author={Chen, Zhennan and Zhu, Junwei and Chen, Xu and Zhang, Jiangning and Chen, Jiawei and Zeng, Zhuoqi and Zhang, Wei and Wang, Chengjie and Yang, Jian and Tai, Ying},
  journal={arXiv preprint arXiv:2605.12013},
  year={2026}
}

@book{cover2009elements,
  title     = {Elements of Information Theory},
  author    = {Cover, Thomas M. and Thomas, Joy A.},
  edition   = {2},
  year      = {2009},
  publisher = {Wiley-Interscience}
}

@inproceedings{gong2026cross,
  title={{Cross-modal Representation Learning for Diffusion-generated Image Detection}},
  author={Gong, Tao and Wang, Dayong and Chu, Qi and Liu, Bin and Yu, Nenghai},
  booktitle=cvpr,
  pages={36092--36102},
  year={2026}
}

@inproceedings{han2024proxedit,
  title={{ProxEdit: Improving Tuning-Free Real Image Editing with Proximal Guidance
}},
  author={Han, Ligong and Wen, Song and Chen, Qi and Zhang, Zhixing and Song, Kunpeng and Ren, Mengwei and Gao, Ruijiang and Stathopoulos, Anastasis and He, Xiaoxiao and Chen, Yuxiao and others},
  booktitle=wacv,
  pages={4279--4289},
  year={2024}
}

@article{zeng2026does,
  title={{When Does High-CFG Diffusion Inversion Fail? A Controlled Study of Prompt–Latent Interactions}},
  author={Zeng, Yan and Hosoya, Yusuke and Tran, Huyen TT and Okatani, Takayuki},
  journal={arXiv preprint arXiv:2607.04731},
  year={2026}
}

@inproceedings{choi2026anchor,
  title     = {{Anchor-Regularized Adaptation for Generalizable AI-Generated Image Detection with DINOv3}},
  author    = {Choi, Hyeongjun and Lee, Juhun and Cozzolino, Davide and Verdoliva, Luisa and Woo, Simon S.},
  booktitle = {ACM Multimedia},
  year      = {2026}
}

@inproceedings{deng2009imagenet,
  author={Deng, Jia and Dong, Wei and Socher, Richard and Li, Li-Jia and Kai Li and Li Fei-Fei},
  booktitle=cvpr, 
  title={{ImageNet: A large-scale hierarchical image database}}, 
  year={2009},
  pages={248-255},
}

@article{lee2026ssafe,
  title   = {{SSAFE: Simple and Strong AI-Generated Image Detection via Frozen Vision Encoders}},
  author  = {Lee, Seunghyun and Kim, Byoungkwon and Nam, Jaehyun and Lee, Kyungmin and Shin, Jinwoo},
  journal = {arXiv preprint arXiv:2606.08634v1},
  year    = {2026}
}
\end{document}